\documentclass[11pt,a4paper]{article}
\usepackage[utf8]{inputenc}
\usepackage[T1]{fontenc}
\usepackage{amsmath,amssymb,amsthm}
\usepackage{mathtools}
\usepackage{bm}
\usepackage{booktabs}
\usepackage[margin=2.5cm]{geometry}
\usepackage{hyperref}
\newtheorem{remark}{Remark}

\DeclareMathOperator*{\argmax}{arg\,max}
\DeclareMathOperator*{\argmin}{arg\,min}
\newcommand{\V}{\mathcal{V}}
\newcommand{\Vbar}{\overline{\mathcal{V}}}
\newcommand{\Pstar}{P^{\star}}
\newcommand{\Ptheta}{P_{\theta}}
\newcommand{\Phat}{P_{\hat{\theta}}}
\newcommand{\Pemp}{\widehat{P}_{N}}
\newcommand{\E}{\mathbb{E}}
\newcommand{\Prob}{\mathbb{P}}
\newcommand{\R}{\mathbb{R}}

\title{The Probabilistic Structure of Large Language Models}
\author{Adnan Aboulalaâ}
\date{}

\begin{document}
\maketitle
\begin{abstract}
This paper presents a probabilistic perspective on large
language models (LLMs), developed with the aim of bringing together, in a single
self-contained account, tools that are usually treated separately across the
literature. LLMs are described through probability measures
on the set of sequences of tokens, specified via their autoregressive conditional
distributions. Training is formulated as a maximum-likelihood estimation problem, addressed by stochastic
gradient methods, while text generation is viewed as the
sequential simulation of the resulting stochastic process. The role of the asymmetry of the
Kullback--Leibler divergence in text generation is examined in relation with characteristic phenomena such as
hallucination and the distinction between statistical plausibility and truth.
As a complementary illustration of the same viewpoint, we also discuss
diffusion models, built around the score function, which cast generation not as sequential 
token prediction but as the simulation of a reverse-time stochastic process transforming noise into
data both in discrete and continuous time.
\end{abstract}

\section{Introduction}

Generative artificial intelligence has moved, within a few years, from a
research topic to an instrument of work. Systems that produce text,
code, images and speech are now used daily across science, engineering,
education, medicine and in many service and industrial sectors ; the economic and social
consequences of this diffusion are already substantial and still unfolding.

\noindent
Generative AI encompasses different probabilistic paradigms, and two families of 
models account for most of this activity: on one hand, autoregressive language models, 
which generate text and code one token at a time via sequential sampling from 
conditional distributions; on the other hand, diffusion models, which generate images, 
audio and video via the reverse simulation of a stochastic process transforming 
noise into data. This paper is concerned with that first family.
A later section extends the discussion to diffusion models, developed in enough depth
to exhibit their own probabilistic machinery --- score functions, stochastic
differential equations, and reverse-time simulation --- so as to illustrate,
from a different generative paradigm, the same interplay between a measure to
be estimated and a stochastic process to be simulated. 

\medskip
At its core, the LLM concept admits a natural formulation within the framework of
probability theory. Let $\V$ denote a finite vocabulary and $\V^{*}$ the set of
finite sequences of tokens over $\V$. A language model defines a probability
distribution on such sequences and, in the autoregressive setting, is
characterized by conditional probabilities of the form
\[
  \Prob\!\left(X_t=w_t\mid X_{<t}=w_{<t}\right),
\]
where $X_t$ denotes the token generated at position $t$ and $X_{<t}$ its
preceding context. A corpus $\mathcal{D}$ provides an empirical distribution
$\Pemp$, while an underlying, generally unknown distribution $\Pstar$ may be
regarded as the data-generating distribution. A parametric family
$\{\Ptheta:\theta\in\Theta\}$ is then introduced to model this distribution. At the
sequence level, the autoregressive factorization reads
\[
  \Ptheta(w_1,\ldots,w_T)=\prod_{t=1}^{T}\Ptheta(w_t\mid w_1,\ldots,w_{t-1}).
\]
Training consists, broadly speaking, in selecting parameters $\theta$ so that
$\Ptheta$ provides a good fit to the empirical distribution $\Pemp$, with the
underlying objective of approximating the unknown data-generating distribution
$\Pstar$.

This probabilistic formulation provides a natural framework for understanding
both the training and the inference processes of large language models. It also
highlights a fundamental distinction between statistical plausibility and truth:
the model assigns likelihood to possible sequences or continuations according to
statistical regularities learned from data, but no term in its objective refers
to the world, and nothing in it represents whether a generated statement is in
fact correct.

\medskip
Stating this framework is easy; stating it precisely is less so. A reader coming
from probability or statistics, and wishing to know exactly \emph{which} measure,
on \emph{which} space, estimated in \emph{which} sense and sampled by
\emph{which} rule, will find the answer harder to assemble than one might
expect. The difficulty is not a shortage of expositions but the shape of the two
literatures that could supply one. Accounts written for a machine-learning
audience are concise and operational, but they routinely leave the central
objects unseparated: the unknown law of the language, the empirical law of the
corpus and the parametric family are all written $P$, and the training criterion
is described as minimizing a ``distance'' whose direction is left unspecified ---
although, as we shall see, that direction governs the behavior of the fitted
model in its tail, and with it the model's propensity to hallucinate. Accounts
written for a probabilistic audience are rigorous but partial: the ingredients
--- autoregressive processes on a finite alphabet, the asymptotic equipartition
property, exponential tilting of a measure, decision rules under a loss --- are
each treated somewhere in the classical literature, but to the best of the
author's knowledge they have not been assembled into a single account of what a
language model is, as a probabilistic object.

\medskip
This paper attempts that assembly. It is written for an expository purpose, 
and it retains the marks of the quest of understanding the machinery through the questions
that actually cause difficulties ; and these questions 
are raised where they arise rather than where a systematic treatment would place
them. Why not use the empirical distribution, since it is already known? What is
missing from the expansion of the divergence? In what sense is the generated text
a Markov chain, given that the process is plainly not memoryless? Once the model
has produced a measure on the vocabulary, how is one word actually chosen from
it? Each of these is elementary, and each is passed over in most presentations. 
More subtle points are addressed, in particular, why the word of the best probability 
is not the one chosen from a probabilistic standpoint and why it is not desirable to be chosen at all.
Concerning these issues and others, the exposition privileges transparency over concision.

\medskip
For completeness on the probabilistic aspects of generative AI, both LLMs and diffusion models are treated in this article; 
the emphasis nonetheless differs between the two: for diffusion models, whose probabilistic
reading is already the standard one and calls for no defense, what is offered
here is essentially depth and completeness of exposition; for language
models, disentangling the relevant probabilistic objects is itself the point
of the exercise, which is why they give the paper its title and its guiding
thread.
For accounts on concepts and techniques of generative AI and LLMs we refer to \cite{vaswani2017}, \cite{AJ},
\cite{RHSSAB}, \cite{AG}, \cite{BCS}, and for a concise presentation, see \cite{AAIgen}.

\medskip
The present paper follows the three stages a language model passes through.
Sections~\ref{section-goal} and \ref{section-model} set up the object: a measure on
sequences, specified by its predictive kernels, together with the parametric
family that approximates it. Section~\ref{section-learning} treats estimation --- the
criterion, its asymmetry, and the algorithms that minimize it.
Section~\ref{section-decoding} treats generation, which we develop at greater length
than is customary, since it is where the probabilistic content is densest and
where expositions are most elliptical; we have in particular tried to make
explicit the two distinct operations that choosing a word involves, the
transformation of the measure and the draw from it, the second of which is
usually passed over in silence. Section~\ref{section-summary} asks in what sense, and
at which levels, the whole enterprise deserves to be called probabilistic, and
where the adjective misleads. Section~\ref{section-diffusion} extends the
discussion to diffusion models, as a complementary illustration: starting from
the discrete-time Gaussian construction, it introduces the score function and
its connection to denoising, and ends with the continuous-time formulation in
terms of stochastic differential equations.

\section{A fundamental goal: construction of a probability measure on sequences of words}
\label{section-goal}

By its very definition, a large language model is a language $L$ endowed with a
probability measure. The measure assigns to every finite sequence of words a
number quantifying how likely that sequence is to occur as a passage of $L$; by
conditioning, it delivers the object of practical interest, the law of the next
word given those already written. Building an LLM amounts to finding a good such
measure by exploring a large corpus and extracting its probabilistic structure,
in three steps:

\begin{enumerate}
\itemsep0pt
  \item fix a large corpus of text in $L$;
  \item train a machine learning system --- transformer-based, this architecture
        having proved the most effective --- to find the probability measure
        that best governs the behavior of word sequences in $L$;
  \item use the resulting probabilistic system for the desired tasks:
        translation, text generation, summarization, question answering.
\end{enumerate}

The sections below will make these steps precise and clarify the probabilistic nature of Gen AI.
\medskip
Fix a finite vocabulary $\V=\{w_1,\dots,w_{|\V|}\}$ (in practice, a set of
\emph{tokens}: sub-word units produced by a byte-pair-encoding scheme
\cite{GPT2}, with
$|\V|$ of the order of $10^{4}$--$10^{5}$).

A sequence of words to be learned or generated is described by $X_1,X_2,\dots:\Omega\to\V$, which are random variables defined on 
a probability space $(\Omega,\mathcal{F},\Prob)$, and their values being in $\V$. Then, the object we wish to model is the law of this process, i.e.\ the family of
finite-dimensional distributions
\begin{equation}
  \Pstar(w_{i_1},\dots,w_{i_m})
  \;:=\;
  \Prob\!\left(X_1=w_{i_1},\dots,X_m=w_{i_m}\right),
  \qquad m\ge 1 .
  \label{eq:fdd}
\end{equation}

\begin{remark}[What $\Pstar$ is]
$\Pstar$ is the law of a passage drawn at random from the totality of text
produced in $L$; equivalently, $\Pstar(s)$ is the relative frequency of the
sequence $s$ among all passages of its length. It is \emph{not} an indicator of
grammaticality: an ungrammatical sequence has small probability, not zero. High
probability therefore means \emph{typical}, not \emph{correct}, cf. Section~\ref{section-summary} below.
\end{remark}

\noindent
{\it Technical note.}
To obtain a probability measure on sequences of finite but unbounded length, we
augment the vocabulary with an absorbing symbol,
$\Vbar=\V\cup\{\texttt{<eos>}\}$, take $\Omega=\bigcup_{m\ge0}\Vbar^{m}$ with the
$X_t$ the coordinate maps, and define the length as the stopping time
$\tau=\inf\{t\ge1:X_t=\texttt{<eos>}\}$. The construction yields a probability
measure on finite strings if and only if $\tau<\infty$ almost surely; a model
violating this leaks mass to infinite sequences and generates without
terminating. Models satisfying $\tau<\infty$ a.s.\ are called \emph{tight}; a
measure-theoretic treatment of this condition, and a proof that the usual model
families are tight, is given in \cite{du2023}.

\noindent
\paragraph{Why considering these probabilities: the predictive objective ?}

The quantity of operational interest is the \emph{one-step predictive
distribution}: given a past of $t-1$ words, the conditional law of the next one,
\begin{equation}
  \Prob\!\left(X_t=\cdot \mid X_1=w_{i_1},\dots,X_{t-1}=w_{i_{t-1}}\right)
  \;=:\; \Pstar\!\left(\cdot \mid w_{<t}\right),
  \label{eq:pred}
\end{equation}
where we write $w_{<t}=(w_{i_1},\dots,w_{i_{t-1}})$ with the usual abuse of
notation. For each fixed context $w_{<t}$, expression \eqref{eq:pred} is a
probability measure on the finite set $\V$.

The joint law and the family of predictive kernels determine each other. In one
direction, the chain rule gives
\begin{equation}
  \Pstar(w_{i_1},\dots,w_{i_m})
  \;=\;
  \prod_{t=1}^{m}\Pstar\!\left(w_{i_t}\mid w_{i_1},\dots,w_{i_{t-1}}\right),
  \label{eq:chain}
\end{equation}
with the convention that the $t=1$ factor is the marginal $\Pstar(w_{i_1})$.

\section{The statistical model}
\label{section-model}

We introduce a parametric family of probability measures:
\begin{equation}
  \mathcal{P}_\Theta=\left\{\Ptheta(\cdot\mid w_{<t}) : \theta\in\Theta\subset\R^{p}\right\},
\end{equation}
realized by a {\it transformer}: the ideal goal is to determine $\Pstar$, but we can not. So the idea is to find a probability measure
$\Ptheta$ that is quite close to $\Pstar$ which governs the probabilities of sequences of words in the language $L$, given what we have observed in the corpus or data set considered. One should
note that $\Pstar\notin\mathcal{P}_\Theta$ in general: the family is
misspecified, and $\hat\theta$ converges not to $\Pstar$ but to the
\emph{pseudo-true} parameter $\argmin_\theta D(\Pstar\|\Ptheta)$, the
information projection of $\Pstar$ onto the family \cite{white1982}.
\\
\noindent
Note. Transformers \cite{vaswani2017} form the AI architecture behind the Gen AI move. They should be viewed as computing systems
whose outputs depend on both the inputs and a huge set of parameters $\theta$. Learning amounts to adjust these parameters.
We refer the reader to \cite{vaswani2017,RHSSAB,AJ,AG,BCS,AAIgen} for these topics.
\\
\\
\noindent
The practical result obtained will be the predictive or conditional probabilities on contexts $w_{<t}$. It is convenient to consider $\Ptheta$
of the form :
\begin{equation}
  \Ptheta(w_t=v\mid w_{<t})
  =\frac{\exp\!\big(z_{t,v}\big)}{\sum_{u\in\V}\exp\!\big(z_{t,u}\big)}
  \;=\;softmax\big(z_t\big)_v,
  \label{eq:softmax}
\end{equation}
(exponential family). The $z_{t,v}=z_t(\theta,w_{<t})\in\R^{|\V|}$ are the so-called {\it logits}. Here $p$, the dimension of $\theta$, is of order $10^{9}$--$10^{12}$; GPT-3
\cite{GPT3}, for instance, has $1.75\times10^{11}$ parameters.

\paragraph{Logits usage:} $z_t$ carries one real \emph{score} per candidate token. The reason to seek $z_t$ is that they are unconstrained: the network may output
any vector of $\R^{|\V|}$, positivity and normalization being restored by the
softmax. Parametrizing the simplex directly would require enforcing $p_v\ge0$
and $\sum_v p_v=1$ at every step, which is awkward.

Taking logarithms in \eqref{eq:softmax}, $\log\Ptheta(v\mid w_{<t})=z_{t,v}-\log\sum_u e^{z_{t,u}}$: the logits are the
log-probabilities up to one additive constant common to all tokens. They are
accordingly defined only up to a shift, $z_t$ and $z_t+c\mathbf 1$ giving the
same measure.

\emph{How the logits are obtained ?} The network compresses the context into a single
vector $h_t\in\R^{d}$ with $d\sim10^{3}$--$10^{4}$, and each token $v$ carries
its own vector $u_v\in\R^{d}$; then
\begin{equation}
  z_{t,v}=\langle h_t,u_v\rangle .
\end{equation}
The score of a candidate is thus the inner product between a summary of the
context and a representation of the token, both living in the same $d$-dimensional
space. Section~\ref{section-transformer-procedure} describes how $h_t$ is built; the essential consequence is
already visible here, and is taken up in Question~1: since contexts with similar
continuations are mapped to nearby $h_t$, an observation made in one context
transfers to contexts never observed \cite{bengio2003}.
\section{The learning phase}
\label{section-learning}

\subsection{Setting}

The corpus that serves for the learning phase is written as: $\mathcal{D}=\{x^{(1)},\dots,x^{(N)}\}$. The $x^{(i)}$ are to be viewed as $N$
realizations drawn (idealized as i.i.d.) from $\Pstar$. Then the following 3 distinct objects are considered:
\begin{itemize}
    \item $\Pstar$ : the true law, unknown --- the \emph{target}.
    \item $\Pemp$ : the empirical law which is observed, but not suitable (see below):
\begin{equation}
  \Pemp=\frac{1}{N}\sum_{i=1}^{N}\delta_{x^{(i)}}
\end{equation}
be the empirical measure on $\V^{*}$. 
		\item $\Ptheta$ : the family of probabilities depending on (huge) set of parameters $\theta$ which are computable, and from which we have to choose 
		the best one that approaches the target $\Pstar$. 
\end{itemize}		

\begin{remark}
The $x^{(i)}$ are \emph{not} copies of the corpus. Taking Wikipedia for example, the articles are
tokenized and concatenated into one long stream, which is then cut into
consecutive blocks of fixed length $L$ --- the context window, say $4096$
tokens. Each block is one $x^{(i)}$, and $N\approx(\text{total tokens})/L$. Long
articles thus span several blocks, short ones share a block with their
neighbors, separated by a document delimiter. The i.i.d.\ idealization is
correspondingly rough: blocks cut from the same article are dependent, and the
cutting points are arbitrary.
\end{remark}

\subsection{The estimation criterion}

We choose $\theta$ so as to minimize some ``distance'' $d(\Pemp,\Ptheta)$ between $\Pemp$ and $\Ptheta$ 
\begin{equation}
\label{eq:klmin}
  \hat{\theta}=\argmin_{\theta\in\Theta}\;d(\Pemp \,\|\, \Ptheta).
\end{equation}
We take :
\begin{equation}
  d(\Pemp,\Ptheta) = D(P\|Q)=\sum_x P(x)\log\frac{P(x)}{Q(x)}.
\end{equation}
$D(P\|Q)$ is the Kullback--Leibler divergence (sometimes written as $KL(P\|Q)$) which is to be viewed here as a sort of distance between the two probabilities $P,Q$. $D(P\|Q)\ge 0$ with equality if and only if $P=Q$. This choice is not a
convention of convenience: minimizing $D(\Pemp\|\Ptheta)$ is equivalent to scoring
the model by the logarithmic loss, which is, up to affine transformation, the
unique \emph{local} strictly proper scoring rule \cite{gneiting2007} ---
``local'' meaning that the score depends only on the probability assigned to the
event actually observed, which is precisely the property allowing the criterion
to decompose along the chain rule \eqref{eq:chain} without a partition function.
Let $H(P)$ be the entropy of $P$:
\begin{equation}
  H(P)=-\sum_x P(x)\log P(x),
\end{equation}
then we have:
\begin{equation}
  D\!\left(\Pemp\|\Ptheta\right)
  = \underbrace{-H(\Pemp)}_{\text{independent of }\theta}
    \;-\;\frac{1}{N}\sum_{i=1}^{N}\log \Ptheta\!\left(x^{(i)}\right),
  \label{eq:expand}
\end{equation}
so that \eqref{eq:klmin} is \emph{exactly} maximum likelihood. Applying the
chain rule \eqref{eq:chain} inside the logarithm yields the loss actually
implemented,
\begin{equation}
  \mathcal{L}(\theta)
  = -\frac{1}{N}\sum_{i=1}^{N}\sum_{t=1}^{T_i}
     \log \Ptheta\!\left(x^{(i)}_t \mid x^{(i)}_{<t}\right).
  \label{eq:loss}
\end{equation}

Equivalently, by the chain rule for the KL divergence,
\begin{equation}
  D\!\left(\Pemp\|\Ptheta\right)
  =\sum_{t\ge1}\E_{\Pemp}\!\left[\,
    D\!\left(\Pemp(\cdot\mid X_{<t})\,\big\|\,\Ptheta(\cdot\mid X_{<t})\right)\right],
  \label{eq:klchain}
\end{equation}
which makes explicit that fitting the joint law is the same as fitting all the
predictive kernels simultaneously, weighted by the frequency of their contexts.

\subsection*{Why not simply using $\Pemp$?} The reader may wonder why not using $\Pemp$ which is already known. 
The main reason is that $\Pemp$ covers only the corpus already known: it will not enable to generate new contents, 
for any new content will have zero probability under $\Pemp$. Even if we smooth it, $\Pemp$ suffers the curse of dimensionality : 
the number of possible sequences grows exponentially with the length $m (|{\cal V}|^m)$. The training set covers only an infinitesimally small fraction of this space.
The parametric family escapes this because it shares statistical strength across
contexts: semantically similar contexts having nearby representations, an
observation transfers to contexts never seen, which a lookup table cannot do
\cite{bengio2003}.

\subsection*{The minimization procedure for (\ref{eq:klmin}):}
We have seen that the learning phase amounts to finding parameters
\[
\hat{\theta}
=
\arg\min_{\theta}\mathcal L(\theta),
\]
where \(\mathcal L(\theta)\) measures how far the probabilities predicted by the
model are from those observed in the training data. For a large language model,
\(\theta\) may contain billions of parameters and \(\mathcal L\) is a highly
non-linear and non-convex function. A direct minimization is therefore impossible
in practice.

The basic tool is \emph{gradient descent}. Starting from an initial value
\(\theta_0\), the parameters are progressively modified in the direction in which
the loss decreases:
\[
\theta_{k+1}
=
\theta_k-\eta_k\nabla_\theta\mathcal L(\theta_k),
\]
where \(\nabla_\theta\mathcal L\) is the gradient of the loss and
\(\eta_k>0\), called the \emph{learning rate}, determines the size of the
modification. The intuition is the same as that of neural networks (explained above): the gradient points locally in the direction of
greatest increase of the loss, so that its opposite,
\[
-\nabla_\theta\mathcal L,
\]
provides a direction in which the loss should decrease.

{\bf Why stochastic gradient descent?} There is, however, a major practical difficulty. The training loss is an average
over an extremely large number \(N\) of training examples:
\[
\mathcal L(\theta)
=
\frac{1}{N}\sum_{i=1}^{N}\ell_i(\theta),
\]
where \(\ell_i(\theta)\) is the loss associated with the \(i\)-th training
example. Consequently,
\[
\nabla_\theta\mathcal L(\theta)
=
\frac{1}{N}
\sum_{i=1}^{N}
\nabla_\theta\ell_i(\theta).
\]
Computing this exact gradient before every parameter update would require
processing the entire training corpus, which would be prohibitively expensive. 
Instead, at each iteration one selects a relatively small random subset of training examples,
\[
B_k\subset\{1,\ldots,N\},
\]
called a \emph{mini-batch}. For example, instead of computing the gradient from billions of training tokens,
one computes it from a much smaller group of examples and defines
\[
\nabla_\theta\mathcal L_{B_k}(\theta)
=
\frac{1}{|B_k|}
\sum_{i\in B_k}
\nabla_\theta\ell_i(\theta).
\]
The parameters are then updated according to
\[
\theta_{k+1}
=
\theta_k
-
\eta_k
\nabla_\theta\mathcal L_{B_k}(\theta_k).
\]
Because the mini-batch is randomly selected, this gradient is random, hence the
terminology \emph{stochastic gradient descent} (SGD). If the examples are sampled uniformly, then
\[
\mathbb E
\left[
\nabla_\theta\mathcal L_{B_k}(\theta)
\right]
=
\nabla_\theta\mathcal L(\theta).
\]

Thus the mini-batch gradient fluctuates around the true gradient, but is much
cheaper to compute. The resulting recursion is a \emph{stochastic approximation}
scheme in the sense of Robbins and Monro \cite{robbins1951}, whose classical
step-size conditions $\sum_k\eta_k=\infty$, $\sum_k\eta_k^{2}<\infty$ are
relaxed in practice to engineered schedules. The basic learning procedure can therefore be summarized as follows:

Select a small part of the training data  $\rightarrow$ Compute the model predictions $\rightarrow$ compute the corresponding loss  $\rightarrow$ compute its gradient $\rightarrow$  slightly modify  $\rightarrow$ select another group of examples and repeat $\ldots$

The entire corpus is therefore progressively processed, but the parameters are
updated many times during this process, rather than only once after processing
the complete corpus.
\noindent
\paragraph{How is the gradient computed?} There remains an apparently big problem. If
\[
\theta=(\theta_1,\ldots,\theta_p),
\qquad p\sim 10^{11},
\]
then the gradient contains
\[
\nabla_\theta\mathcal L
=
\left(
\frac{\partial\mathcal L}{\partial\theta_1},
\ldots,
\frac{\partial\mathcal L}{\partial\theta_p}
\right),
\]
possibly hundreds of billions of partial derivatives.These derivatives are not computed separately. As we saw, neural networks use
\emph{backpropagation}, which is an efficient application of the chain rule of differentiation
\paragraph{From SGD to Adam.} The elementary SGD update
\[
\theta_{k+1}
=
\theta_k-\eta_k g_k,
\qquad
g_k=\nabla_\theta\mathcal L_{B_k}(\theta_k),
\]
uses essentially the same learning-rate factor \(\eta_k\) for every parameter. 
The learning rate \(\eta_k\) is itself usually varied during training.

In a large neural network, however, different parameters can have gradients with
very different magnitudes and behaviors. Modern LLM training therefore commonly
uses adaptive variants of stochastic gradient descent, particularly \emph{Adam}
and \emph{AdamW}. Adam was introduced by Kingma and Ba~\cite{kingma2015}, and the decoupled
weight-decay version AdamW by Loshchilov and Hutter~\cite{loshchilov2019}. We omit the details.

\paragraph{Remark on the parameters $\theta$.}
While the numerical optimization procedure determines a parameter vector
$\widehat{\theta}$, the object of interest is ultimately the probability model
$P_{\widehat{\theta}}$ defined by these parameters. The parameter vector itself
is generally not uniquely identifiable. Indeed, because of symmetries in neural
network parameterizations (such as permutations of hidden units or certain
compensating rescalings), two different parameter vectors may define the same,
or essentially the same, probability model:
\[
    \theta \neq \theta'
    \qquad\text{while possibly}\qquad
    P_{\theta}=P_{\theta'}.
\]
Thus, one should not interpret training as recovering a unique ``true''
parameter vector. Moreover, since the loss function is highly non-convex and
the optimization is performed over a finite number of iterations, the training
procedure is not guaranteed to reach a global minimizer. The resulting
$\widehat{\theta}$ should rather be viewed as one parameter configuration
reached by the optimization procedure and providing a satisfactory fitted
probability model.

Once training is completed, the parameters $\widehat{\theta}$ are fixed.
For a given context $w_{<t}$, the Transformer uses them to compute the
conditional probability distribution
\[
    P_{\widehat{\theta}}
    \bigl(\,\cdot\mid w_{<t}\bigr).
\]
The collection of these conditional distributions defines the learned
autoregressive probability model
\[
    P_{\widehat{\theta}}(w_{1:T})
    =
    \prod_{t=1}^{T}
    P_{\widehat{\theta}}(w_t\mid w_{<t}).
\]
Hence, $\widehat{\theta}$ is primarily an internal numerical
parameterization: the probabilistic object of interest is the model
$P_{\widehat{\theta}}$ that it defines.

To end this paragraph let us recall, the complete learning process :
\\
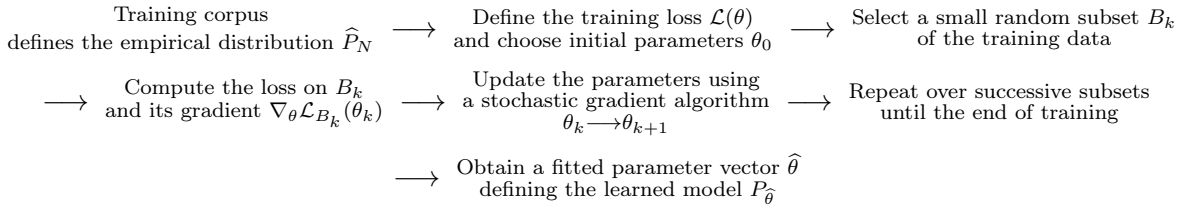
\[
\substack{\text{Training corpus}\\
          \text{defines the empirical distribution } \widehat{P}_{N}}
\;\longrightarrow\;
\substack{\text{Define the training loss }\mathcal L(\theta)\\
          \text{and choose initial parameters }\theta_0}
\;\longrightarrow\;
\substack{\text{Select a small random subset }B_k\\
          \text{of the training data}}
\]
\[
\;\longrightarrow\;
\substack{\text{Compute the loss on }B_k\\
          \text{and its gradient }
          \nabla_\theta\mathcal L_{B_k}(\theta_k)}
\;\longrightarrow\;
\substack{\text{Update the parameters using}\\
          \text{a stochastic gradient algorithm}\\
          \theta_k\longrightarrow\theta_{k+1}}
\;\longrightarrow\;
\substack{\text{Repeat over successive subsets}\\
          \text{until the end of training}}
\]
\[
\;\longrightarrow\;
\substack{\text{Obtain a fitted parameter vector }\widehat\theta\\
          \text{defining the learned model }P_{\widehat\theta}}
\]

\section{The inference phase or how to generate texts and next words from the probability measure obtained}
\label{section-decoding}

Having obtained $\hat\theta$, we have at our disposal, for every context, the
predictive measure
\begin{equation}
  \pi_t(\cdot) \;:=\; \Phat\!\left(\cdot \mid w_{<t}\right)
  \;\in\; \mathcal{M}_1(\V),
\end{equation}
a probability measure on the vocabulary.

\begin{remark}
It is worth resisting the formulation ``predict the next best word''. The output of
the model is $\pi_t$ in its entirety --- a measure, not a point. There is no
``best'' word supplied by the model; selecting one is a separate \emph{decision}
problem, with a loss function of its own, treated in
this section. Likewise $\Phat$ is close to $\Pemp$ by
construction; whether it is close to $\Pstar$ is the question of
generalization, which is not addressed by \eqref{eq:klmin}.
\end{remark}

\begin{remark}[Technical note on the Markov property of the text generated]
\label{rem:markov}
The transformer has a finite context window $L$: by construction the attention
mask lets position $t$ see at most the $L$ preceding tokens, so that for $t>L$
\begin{equation}
  \Phat\!\left(\cdot\mid w_{<t}\right)=\Phat\!\left(\cdot\mid w_{t-L:t-1}\right).
  \label{eq:orderL}
\end{equation}
The process $(X_t)$ is thus of order $L$, and $X_t$ alone is \emph{not} Markov.
It becomes so after the standard change of state space. Put
\begin{equation}
  Y_t=(X_{t-L+1},\dots,X_t)\in\V^{L},
\end{equation}
the sliding window. Given $Y_t=(x_{t-L+1},\dots,x_t)$, the next state $Y_{t+1}$
is $(x_{t-L+2},\dots,x_t,v)$ with $v=X_{t+1}$: the first $L-1$ coordinates are
a deterministic shift of $Y_t$, and only the last is random, drawn according to
\eqref{eq:orderL}. Hence
\begin{equation}
  \Prob\!\left(Y_{t+1}=y'\mid Y_t=y,\,Y_{t-1},\dots\right)
  =\begin{cases}
     \Phat(v\mid y) & \text{if } y'=(y_{2},\dots,y_{L},v),\\[2pt]
     0 & \text{otherwise,}
   \end{cases}
\end{equation}
which depends on the past only through $Y_t$: $(Y_t)_{t>L}$ is a Markov chain on
$\V^{L}$, and generation is exactly its simulation, started from the prompt.
\end{remark}

\paragraph{Decoding: from a probability measure to a word}

At each step the model hands us a probability measure $\pi_t$ on $\V$, not a word.
Producing one decomposes into two stages: \textbf{(a)} a transformation of the
measure, $\pi_t\mapsto\tilde\pi_t$, still a probability on $\V$: the aim here is to optimize further the measure $\pi_t$ which will guide the choice of the next word : for instance by restricting it to some of the best candidates; \textbf{(b)}
the realization of one draw from $\tilde\pi_t$. Stage (a) distinguishes the
strategies; stage (b) is common to all and is where the randomness enters.

\subsection{A visual analogy of the decoding process}

Suppose the measure has been restricted to 20 words candidates. Cut a segment of
length $1$ into twenty pieces, the length of each being the probability of its
word: a word of probability $0.45$ occupies $45\%$ of the segment. Throw a dart
randomly (and uniformly) at the segment; the word whose piece is hit is the next word to be generated.

The most probable word is not necessarily the \emph{chosen} one, although it
owns the largest piece and is therefore hit most often. 
Among the choice strategies that will be reviewed below, let us mention : \emph{Greedy decoding}
throws no dart and takes the largest piece ; \emph{Truncation} deletes the
minuscule pieces before throwing and stretches the rest back to unit length ---
the renormalization ;  \emph{Temperature $T$} rescales the pieces before throwing: with low $T$
(cold), the large pieces grow and the small shrink; with high $T$ (hot), they tend to equalize. Truncation and
temperature alter the sizes of the pieces.

\subsection{Stage (a): transforming the measure}

Write $\pi=\pi_t$, with logits $z$, $\pi=softmax(z)$.

\paragraph{Greedy decoding : } $X_t=\argmax_v\pi(v)$; It is deterministic. It maximizes the
probability of the \emph{next} token only, not of the sequence, since the
maximum does not factorize into a product of individual measures. It is a myopic choice for it considers only what is the right word for the next step (only) and not for 
whole text to be generated. It produces flat text (lack of diversity) and falls into repetitive loops (``I see I see ...'') because once a high-probability token is chosen, 
the context shifts to reinforce the same choice.

\paragraph{Beam search :} it is a deterministic strategy that approximates the global mode of the sequence:
\[
\hat{w}_{1:T} = \arg\max_{w_{1:T}} P_{\hat{\theta}}(w_{1:T}),
\]

or more commonly, a length-normalized variant \(T^{-\alpha}\log P_{\hat{\theta}}(w_{1:T})\). At each step, it maintains a fixed number \(k\) (the beam width) of 
the most probable partial sequences, extends each by one token, and keeps the \(k\) best among the resulting \(k|\mathcal{V}|\) candidates. The advantage is that 
it {\it searches globally} and is therefore excellent for tasks where the output is largely determined by the input, such as machine translation, speech recognition, 
and summarization — precisely the tasks for which it was originally designed.

For open-ended text generation, however, beam search performs poorly. The reason is the {\it likelihood trap}: the mode of a high-dimensional distribution is 
systematically atypical. 
The most probable sequence is the most predictable, hence least informative, and often degenerates into short, repetitive, or bland text. 
Moreover, beam search collapses diversity across different prompts; a fixed beam width produces similar outputs regardless of the prompt's uncertainty. 
Consequently, beam search has been largely abandoned in conversational and creative generation systems as of 2026.

\paragraph{Ancestral (or pure) sampling:} here stage (a) is the identity,
$\tilde\pi=\pi$, and one simply draws $X_t\sim\pi_t$. The name records the
order of the draws: one follows the factorization \eqref{eq:chain} in its
natural direction, drawing $X_1$ from $\pi_1$, then $X_2$ from
$\pi_2(\cdot\mid X_1)$, and so on, each variable being drawn only after its
``ancestors'' in the chain rule. It is the only rule producing an \emph{exact}
realization of the fitted process $\Phat$ --- see \S\ref{section-notgreedy} --- and
is the correct choice if one's aim is to study $\Phat$ itself.

Its weakness in application follows from the mass-covering property established
in Section~\ref{section-learning}. By \eqref{eq:softmax} all
$|\V|\approx5\times10^{4}$ tokens carry strictly positive probability, and the
aggregate mass of the tail is typically a few percent, say $3\%$. Over a
generation of $500$ tokens the probability of drawing at least one implausible
token is then $1-(1-0.03)^{500}\approx1$; and a single such token derails
everything that follows, since the model conditions on it. Hence the truncation
schemes below.

\paragraph{Temperature Scaling or Exponential Tilting:} temperature scaling modifies the predictive measure by raising the logits (or equivalently, exponentiating the probabilities) 
with a temperature parameter \(T > 0\):

\[
\pi^{(T)}(v) = \frac{\exp(z_v / T)}{\sum_{u \in \mathcal{V}} \exp(z_u / T)} \propto \pi(v)^{1/T}.
\]
This is an exponential tilting of the original distribution: it is a Gibbs measure with the log-likelihood as energy and \(T\) as the inverse temperature. As \(T \to 0^+\), 
the distribution concentrates on the mode (recovering greedy decoding in the limit); as \(T = 1\), it leaves \(\pi\) unchanged; as \(T \to \infty\), 
it flattens to the uniform measure. The entropy \(H(\pi^{(T)})\) increases monotonically with \(T\).

The advantage is {\it fine-grained control} over the randomness of the output: Lower temperatures (e.g., \(T = 0.6\)) reduce the probability of rare tokens, 
producing more focused and deterministic text; higher temperatures (e.g., \(T = 1.2\)) increase diversity and surprise. The disadvantage is that temperature 
{\it does not remove the tail} of the distribution: every token, no matter how improbable, retains strictly positive probability. In practice, it is almost always 
combined with truncation (top-\(k\) or top-\(p\)) to avoid drawing from the inflated tail. Furthermore, optimal \(T\) is context-dependent and requires manual tuning.

\paragraph{Top-$k$ sampling (fixed truncation) :} this sampling mode restricts the support to the \(k\) most probable tokens, renormalises, and then samples from the 
truncated distribution:

\[
\tilde{\pi}(v) = \frac{\pi(v)}{\sum_{u \in A_k} \pi(u)} \quad \text{for } v \in A_k, \quad A_k = \text{the } k \text{ most probable tokens}.
\]

The advantage is that it removes the long, noisy tail of the distribution, drastically reducing the chance of sampling rare or nonsensical tokens. It is more diverse 
than greedy and straightforward to implement. The disadvantage is that \(k\) is \textbf{fixed}, while the entropy of \(\pi_t\) varies dramatically across contexts. 
At a highly predictable position (e.g., after ``The capital of France is''), the distribution is peaked; a fixed \(k\) keeps \(k-1\) implausible candidates. At a highly 
uncertain position (e.g., after ``The story begins with''), the distribution is flat; the same \(k\) may discard legitimate alternatives. 
This rigidity is the primary motivation for the adaptive strategy that follows.

\paragraph{Top-$p$ (nucleus or dynamic truncation):} also known as top-\(p\) sampling, it is an improvement of Top-$k$ method, by allowing variations of $k$: 
it dynamically selects the smallest set \(A_p\) of tokens whose cumulative probability exceeds a threshold \(p \in [0.9, 0.95]\):

\[
A_p = \arg\min_{A} \left\{ \pi(A) \geq p \right\}, \qquad \tilde{\pi}(v) = \frac{\pi(v)}{\pi(A_p)} \quad \text{for } v \in A_p.
\]

Introduced in \cite{holtzman2020}, the set \(A_p\) is the ``nucleus'' of the distribution: the minimal set that carries a fraction \(p\) of the mass. 
The advantage is that \(|A_p|\) adapts to the entropy of \(\pi_t\). When the model is confident, the nucleus is small (sometimes just 1 or 2 tokens); when uncertain, 
the nucleus is large. This adaptivity makes it the default in deployed conversational systems (OpenAI, Anthropic, Google). The disadvantage is that the threshold \(p\) 
itself still requires tuning, and the procedure may include rare tokens if the probability mass is smoothly distributed. Nevertheless, 
it has largely superseded top-\(k\) in practice, and is typically combined with temperature scaling (order: temperature, then nucleus conditioning).

\paragraph{Typical sampling (information-theoretic truncation):} introduced by Meister \emph{et al.}~\cite{meister2023}, this is a more recent information-theoretic strategy. 
Instead of retaining the most probable tokens, it retains tokens whose negative log-probability \(-\log \pi(v)\) is close to the entropy \(H(\pi)\) of the distribution. 
The intuition comes from the asymptotic equipartition property \cite{cover2006}: typical sequences are those whose log-probability is near the entropy rate. Specifically, 
one selects a set

\[
A_{\text{typical}} = \left\{ v : \left| -\log \pi(v) - H(\pi) \right| \leq \delta \right\},
\]

for some threshold \(\delta\), and then samples from the renormalized distribution over \(A_{\text{typical}}\). The advantage is that it deliberately \textbf{avoids the mode} 
(see the note below), which is systematically atypical, and produces text that is more representative of the model's ``typical'' outputs rather than its ``most likely'' ones. 
This can reduce repetitive and dull generations. The disadvantage is that the threshold \(\delta\) is non-trivial to tune, and the method is less widely adopted than nucleus sampling 
due to its relative novelty and conceptual complexity.

\subsection{The mode, and why it is avoided ?} The {\it mode} of a probability
measure is its most probable outcome: $\argmax_{v}\pi(v)$ for a single token ---
the word greedy decoding selects --- and $\argmax_{w_{1:T}}\Phat(w_{1:T})$ for a
whole text, which beam search approximates. Two distinct points must be kept
apart here. The first is probabilistic and explains why a draw does not produce
the mode; the second is semantic and explains why one would not want it even if
it did.

\medskip
\emph{First point: being the most probable at each step confers almost no
advantage over many steps.} Return to the measure of \S\,4.1, whose leading word
carries probability $0.45$. Greedy decoding selects it at every step. Over $7$
steps the modal path keeps its rank --- no other particular sequence is more
probable --- and yet its probability is only $0.45^{7}\approx0.0023$, that is,
fewer than one draw in $400$; over $10$ steps, one in $5\,900$; over $100$
steps, one in $10^{37}$. A random draw therefore almost never lands exactly on
the mode. In fact

An elementary comparison: let $ X_i$ be a sequence of i.i.d. random variables with values $A=1$ (head) 
with probability $p_A= 0.9$ and $B=-1$ (tails) with probability $p_{B}=0.1$.
The mean the $X_i$ is $M=A p_A + B p_B= 0.8$ and by the law of large numbers, all the paths 
$ X_1, X_2, … X_n $ whose empirical means $( X_1+ \ldots + X_n)/n$ is {\it very} close to the 
expectation $M=0.8$ are eligible to happen; but the sequence $1 1 1 1 1 1 \ldots $ of the most 
probable outcome (A) at each step, has $1$ as empirical mean, which is not {\it very} close to 
the expectation $M=0.8$. We know already that the sequence $1 1 1 1 1 1 \ldots $ is very unlikely to occur by the 
law of large numbers, and in fact the probability of events that are around this sequence is even 
exponentially small w.r.t $n$.
In other words, in a head and tails game, obtaining a sequence of head in $n$ tosses is very unlikely to 
happen even if the probability to obtain head is $90 \%$ at each step.

\medskip
\emph{Second point: why the mode should be avoided anyway.} This second argument
is not probabilistic but semantic and informational. A text that is maximally
probable at every step is entirely predictable: it maximizes likelihood and
minimizes surprise, and therefore carries almost no information. This is exactly
the flat, repetitive, loop-prone output observed with greedy decoding and beam
search, and it is known as the \emph{likelihood trap} \cite{holtzman2020}.

Human text is neither maximally predictable nor random; it holds a balance
between the two. It sits in what information theory calls the \emph{typical
set}: the region where the probability per token is close to the entropy of the
language \cite{cover2006}. Typical sampling is the decoding rule that targets
that region directly, instead of climbing toward a peak that is both unreachable
by sampling and undesirable as text.

\paragraph{Composition and interpretation.} These compose, conventionally in the
order temperature, top-$k$, top-$p$, renormalization; the order matters and is a
convention. Truncation should be read as a \emph{correction for
misspecification in the tail}.

\subsection{Stage (b): performing the draw or how the next word Is finally chosen}

Once the predictive measure \(\pi_t\) has been transformed into a modified distribution \(\tilde{\pi}_t\) via temperature scaling, truncation (top-\(k\), top-\(p\)), 
or typical set selection, the final step is to {\it realize a single draw} from \(\tilde{\pi}_t\). Except for greedy decoding (which bypasses this step 
by deterministically taking the \(\arg\max\)), all other strategies require a concrete algorithmic procedure to select a token according to the probabilities
 specified by \(\tilde{\pi}_t\). The mechanics of this selection fall into two standard methods: {\it inverse transform sampling} (the conceptual gold standard) 
 and the {\it Gumbel-max trick} (the computationally efficient workhorse).

Let $A=\{v_1,\dots,v_m\}$ be the support of $\tilde\pi$, that is, the set of
tokens still carrying positive probability after stage (a). Thus
$m=|A|=|\V|$ if no truncation was applied, $m=k$ under top-$k$, and
$m=|A_p|$ under top-$p$ --- in which case $m$ varies from step to step, being
small where $\pi_t$ is concentrated and large where it is spread out.

The enumeration $v_1,\dots,v_m$ is merely a labeling of the elements of $A$:
we number them in order to write down cumulative sums. The order is neither
random nor prescribed, and the law of $X_t$ does not depend on it --- the
segment below is cut into the same $m$ pieces whatever the arrangement, and a
uniform dart is indifferent to how they are arranged. Both $A$ and its
labeling are of course recomputed at every step $t$, since $\pi_t$ changes with
the context. In practice the decreasing order is used, because top-$p$ has
already sorted $\pi$ in order to determine $A_p$, so the cumulative sums come
free and the search below is a binary search.

\paragraph{Inverse transform.} With $F_0=0$ and $F_j=\sum_{i\le j}\tilde\pi(v_i)$
for $j=1,\dots,m$, so that $F_m=1$, let $U$ be a random variable uniformly
distributed on $[0,1]$, drawn independently at each step. Given its realization
$U(\omega)$, set
\begin{equation}
  X_t=v_J,\qquad J=\min\{j:\;U(\omega)\le F_j\}.
\end{equation}
The rule is correct because
$\{J=j\}=\{F_{j-1}<U\le F_j\}$, an interval of length $F_j-F_{j-1}$, and a
uniform variable falls in an interval with probability equal to its length;
hence
\begin{equation}
  \Prob(X_t=v_j)=F_j-F_{j-1}=\tilde\pi(v_j).
\end{equation}
This is the dart and the segment, literally: the $F_j$ are the cut points and
$U(\omega)$ is the dart.

Two points deserve emphasis. First, the successive draws $U_1,U_2,\dots$ are
independent of one another and of everything else; this independence is what
makes the sequential procedure produce an exact sample from the joint law, as
shown in \S\ref{section-notgreedy}. Second, all the modeling has already been done
by the time we reach this step: the transformer determines the lengths of the
pieces, and the dart is a plain uniform variable carrying no information about
language.

\paragraph{Gumbel-max.} In practice, particularly on parallel hardware such as GPUs, inverse transform sampling
is not optimal because it requires sorting the tokens or computing cumulative sums sequentially. The Gumbel-max trick provides a mathematically equivalent but 
parallelisable alternative: it avoids both the sorting and the normalization: perturb each
logit by independent noise and take the largest.

Let $G$ follow the standard \emph{Gumbel} distribution,
$\Prob(G\le x)=\exp(-e^{-x})$ --- the classical extreme-value law of type I, and
easy to simulate, since $G=-\log(-\log U)$ with $U$ uniform on $[0,1]$. Draw
$(G_v)_{v\in\V}$ i.i.d.\ from it and set
\begin{equation}
  X_t=\argmax_{v\in\V}\left(\frac{z_v}{T}+G_v\right).
  \label{eq:gumbelmax}
\end{equation}
Then then \(X_t\) is distributed exactly according to \(\tilde{\pi}\) : $X_t\sim\pi^{(T)}$. The verification is easy: Write $s_v=z_v/T$ and
$M_v=s_v+G_v$, so that $\Prob(M_v\le x)=\exp(-e^{s_v}e^{-x})$ and
$M_v$ has density $e^{s_v}e^{-x}\exp(-e^{s_v}e^{-x})$. Then
\begin{equation}
  \Prob\!\left(\argmax_u M_u=v\right)
  =\int_{\R} e^{s_v}e^{-x}
     \exp\!\Big(-\Big(\textstyle\sum_{u}e^{s_u}\Big)e^{-x}\Big)\,dx
  =\frac{e^{s_v}}{\sum_u e^{s_u}}
  =\pi^{(T)}(v),
\end{equation}
the middle equality following from the substitution $y=e^{-x}$, which turns the
integral into $\int_0^\infty e^{s_v}e^{-Sy}dy$ with $S=\sum_u e^{s_u}$.

The Gumbel-max trick is a direct consequence of the fact that the Gumbel distribution is the max-stable distribution for
the exponential family (\ref{eq:gumbelmax}). Its advantage is computational: it requires only one pass over the
vocabulary (to add noise and take a maximum), with no sorting, no cumulative sums,
and no binary search. Every coordinate is treated independently ---
so the whole operation is one parallel pass followed by a maximum --- and that
the normalizing constant never appears. Truncation is incorporated simply by
setting $z_v=-\infty$ for $v\notin A$ before applying \eqref{eq:gumbelmax}. As a
by-product, the maximum itself is Gumbel with location $\log\sum_u e^{s_u}$ and
is independent of the argmax, which is what allows the construction to be
extended to sampling several tokens without replacement. This is why modern implementations of temperature scaling and nucleus sampling use this trick internally; 
it is particularly efficient when the support of \(\tilde{\pi}\) is the full vocabulary and the distribution is evaluated in vectorized operations.

\subsection{Why truncated sampling does not collapse to greedy decoding}
\label{section-notgreedy}

We rephrase here the arguments discussed above on the mode. The most probable word keeps the largest piece under every scheme, so sampling
might seem a mere softening of greedy decoding, inheriting its myopia. But it is
not the case:

\medskip
(1) Greedy decoding takes the most probable word at each step. It is tempting to think that sampling will reproduce this most probable word. But it does not : 
it writes it with probability $0.45$, but writes something else $58\%$ of the time : the advantage of the most probable word is no guarantee that it will be chosen. 
Furthermore, the probability for choosing the most probable words in $T$ steps, is, e.g., $0.45^T$ which will be very small if $T$ is just $10$ or $20$ ; 
here we assume the same probability $0.45$ for the most probable words in the $T$ steps.

\medskip
(2) The aim of not choosing, {\it with certainty}, the most probable next word (i.e. not adopting greedy decoding) is not variety for its own sake. 
If a continuation is correct in $45\%$ of such contexts, writing it $45\%$ of the
time is right; writing it always over-represents it, writing it $10\%$ of the
time may under-represents it. 

\medskip
(3) Greedy decoding is myopic because
\begin{equation}
  \argmax_{w_{1:T}}\prod_{t}\pi_t(w_t)
  \;\neq\;\left(\argmax\pi_1,\dots,\argmax\pi_T\right):
\end{equation}
maximizing stepwise does not maximize the product. Sampling, by contrast,
factorizes \emph{exactly}: drawing $X_1\sim\pi_1$, then $X_2\sim\pi_2(\cdot\mid
X_1)$, and so on, the chain rule \eqref{eq:chain} gives
$(X_1,\dots,X_T)\sim\Phat(w_1,\dots,w_T)$ with no approximation. A purely local
procedure yields an exact sample from the global joint law.

\begin{remark}[The decision-theoretic answer]
Decoding being a decision problem, it has a loss function, and the mode is
simply the rule attached to the $0$--$1$ loss $u(y,y')=\mathbf 1\{y=y'\}$ --- a
loss with no discrimination whatever on a combinatorial space, which is why
maximizing probability is the wrong objective rather than merely an imperfect
one. \emph{Minimum Bayes risk} decoding replaces it by
\begin{equation}
  \hat y=\argmax_{y}\;\E_{y'\sim\Phat}\!\left[u(y,y')\right],
\end{equation}
for a utility $u$ measuring textual similarity, the expectation being estimated
by Monte Carlo from ancestral samples \cite{eikema2020}. The draws of
\S\,6.3 thus become the engine of a decision rule, and not merely the mode of
generation.
\end{remark}

\subsection{Practice in deployed systems}

As of 2026, beam search has disappeared from conversational systems; the norm is
ancestral sampling with temperature and nucleus truncation. The API defaults of
OpenAI, Anthropic and Google all set temperature to $1.0$, and all three
recommend adjusting either temperature or top-$p$, not both --- a recommendation
without theoretical basis, composing a tilting and a conditioning being
perfectly well defined, but hard to tune blind. Open-weight models are the only
case where the settings of a deployed product are documented: DeepSeek reports
$T=0.6$ and $p=0.95$, the former being the value used in its web interface, with
the range $0.5$--$0.7$ recommended to avoid both endless repetition and
incoherence --- precisely the two pathologies of a tilting that is too cold or
too hot. Closed products (ChatGPT, Claude, Gemini) do not disclose theirs. A
recent trend removes the knobs altogether: several frontier models now reject
non-default sampling parameters, the decoder ceasing to be a user choice and
becoming a component of the model, co-tuned with post-training.

\subsection{The mass-covering property of the forward KL divergence}
The mass-covering property, or zero-avoiding property is a behavior stemming from the mathematical asymmetry of the Kullback--Leibler divergence, 
specifically the Forward KL, $D(P || Q)$. The contrast between the forward direction, which is mass-covering (or \emph{inclusive}), 
and the reverse direction $D(Q || P)$, which is mode-seeking (or \emph{exclusive}), is standard; see Bishop~\cite[\S10.1.2]{bishop2006} 
for a graphical account and Minka~\cite{minka2005} for the general treatment within the $\alpha$-divergence family.
\\
\noindent
In the context of training models like LLMs, if $P_{e}$ represents the empirical data distribution of the training corpus (our $\widehat{P}_{N}$), the sum in the formula:
\[
D (P_e || \Ptheta) = \sum_{x} P_e(x) \log \frac{P_e(x)}{\Ptheta(x)}
\]

is weighted by the true data distribution $P_e(x)$. If there is any specific data point or sequence where the empirical distribution has mass ($P_e(x) > 0$), but the model assigns it a probability near zero ($\Ptheta(x) \to 0$), the ratio $P_e(x)/\Ptheta(x)$ grows infinitely large. This drives the divergence, and therefore the training loss, toward $+\infty$.
\\
\noindent
The Result (Mass-Covering): To avoid this infinite penalty, the model $\Ptheta$ is {\it forced} to assign non-zero probability everywhere the true data $P_e$ has mass and this is what happens:

\begin{itemize}
\item Support Inclusion vs. Support Restriction: at a first sight one may think that this will favor the trained corpus, but it is the opposite that happens. The penalty enforces $P_e(x) > 0 \implies \Ptheta(x) > 0$, setting a strict floor on $\Ptheta(x)$ across the training set. Critically, it places zero penalty on $\Ptheta(x)$ being non-zero where $P_e(x) = 0$. It compels $\Ptheta$ to cover the empirical corpus, but never forbids $P_e$ from expanding beyond it.

\item Forced Smoothing and Generalization: Because $\Ptheta$ is parametrized by a continuous Transformer rather than a discrete lookup table, it cannot assign non-zero probability to every training corpus sequence $x \in \mathcal{D}$ without smoothly assigning non-zero probability to the continuous space surrounding them. The requirement to cover all training points forces $\Ptheta$ to "fill in the gaps," assigning probability mass to unseen, newly generated sequences ($x \notin \mathcal{V}^{*}$).

\item By forcing $\Ptheta$ to maintain non-zero probability across all empirical data, Forward KL forces the model to maintain broad, permissive support. This open support is the exact probabilistic mechanism that allows generative models to sample valid, novel token sequences during inference.

\end{itemize}
\section{Summary of the decoder transformer procedure}
\label{section-transformer-procedure}

For completeness we review here the decoder transformer \cite{vaswani2017} procedure which is the part used in Gen AI systems.

In a mathematical form, the decoder transformer can be considered as a map from a token sequence to logits. We describe
its procedure and the learning and inference phases.

\paragraph{Representations.} Let $d$ be the model dimension. To each position
$t$ and each layer $\ell$ is attached a vector $h_t^{(\ell)}\in\R^{d}$, the
\emph{representation} of position $t$ after layer $\ell$: a summary of
$w_{\le t}$, refined layer by layer, from which the logits will eventually be
read off. Stacking them row-wise gives $H^{(\ell)}\in\R^{T\times d}$. With
$E\in\R^{|\V|\times d}$ the embedding matrix and $p_t$ a positional encoding,
\begin{equation}
  h^{(0)}_t = E_{w_t}+p_t .
\end{equation}

\paragraph{Layers.} For $\ell=1,\dots,L_{\mathrm{d}}$,
\begin{align}
  \tilde h^{(\ell)} &= h^{(\ell-1)} + MHA\!\left(LN\big(h^{(\ell-1)}\big)\right),\\
  h^{(\ell)} &= \tilde h^{(\ell)} + FFN\!\left(LN\big(\tilde h^{(\ell)}\big)\right),
\end{align}
where the three operators are:
\begin{itemize}
\itemsep0pt
  \item $LN$ (\emph{layer normalization}) standardizes each vector across its
    $d$ coordinates and rescales it, $LN(h)=\gamma\odot(h-\mu)/\sigma+\beta$
    with $\mu,\sigma$ the mean and standard deviation of the entries of $h$ and
    $\gamma,\beta\in\R^{d}$ learned; it keeps activation scales stable with
    depth.
  \item $MHA$ (\emph{multi-head attention}) is the only operator mixing
    positions. For one head, with $Q=HW_Q$, $K=HW_K$, $V=HW_V$,
    \begin{equation}
      Attn(Q,K,V)=softmax\!\left(\frac{QK^{\top}}{\sqrt{d_k}}+M\right)V,
      \qquad
      M_{ts}=\begin{cases} 0 & s\le t,\\ -\infty & s>t;\end{cases}
      \label{eq:causal}
    \end{equation}
    $MHA$ concatenates several such heads with distinct projections and
    recombines them by $W_O$.
  \item $FFN$ (\emph{feed-forward}) acts on each position separately,
    $FFN(h)=W_2\,\phi(W_1h)$ with $W_1\in\R^{d_f\times d}$,
    $W_2\in\R^{d\times d_f}$, $d_f\approx4d$ and $\phi$ a nonlinearity.
\end{itemize}
Here $\odot$ denotes the componentwise product: for
$a,b\in\R^{d}$, $(a\odot b)_i=a_ib_i$. Thus $\gamma$ rescales each of the $d$
coordinates separately, and $\beta$ shifts each separately, after the vector has
been standardized.

Finally, with $W_U\in\R^{|\V|\times d}$ the unembedding matrix,
\begin{equation}
  z_t = W_U\,LN\!\left(h^{(L_{\mathrm{d}})}_t\right),
  \qquad
  \Ptheta(\cdot\mid w_{\le t})=softmax(z_t),
\end{equation}
which is the inner product $\langle h_t,u_v\rangle$ of Section~\ref{section-model}, the rows of
$W_U$ being the token vectors $u_v$. The parameter is the whole collection
$\theta=\{E,\,p_\cdot,\,W_Q,W_K,W_V,W_O,\,W_1,W_2,\,\gamma,\beta,\,W_U\}$ over
all layers and heads.
\\
\noindent
NB.The causal mask described in the § on the transformers is implemented by $M$ in \eqref{eq:causal} is the architectural
encoding of the chain rule: it forces $z_t$ to depend on $w_{\le t}$ only, so
that the network computes genuinely \emph{predictive} kernels. Without it the
factorization \eqref{eq:chain} would not be satisfied and the model would ``predict''
a token by looking at it.

\paragraph{Learning phase :} this was described in Section~\ref{section-learning}. The above procedure just presented the function being differentiated. The composition
above is smooth in $\theta$, so $\nabla_\theta\mathcal{L}$ is obtained by the
chain rule of calculus applied backwards through the layers --- backpropagation
--- and $\theta$ is updated by \eqref{eq:loss} and the minimization algorithm of $\mathcal{L}$. 
This paragraph provides an important practical fact: thanks to the causal mask, a
\emph{single} forward pass on $w_{1:T}$ produces all $T$ logit vectors in
parallel, hence all $T$ conditionals of \eqref{eq:loss} at once, differentiated
in one backward pass. The contexts used are always the true corpus prefixes,
never the model's own output --- this is \emph{teacher forcing}. Recurrent
architectures cannot do this, which is the practical reason transformers scale.

\paragraph{Inference phase :} the situation is the reverse of the preceding one.
During learning the sequence is known in advance, so every position can be
processed at once. During generation the sequence is being created as we go:
$z_{t+1}$ cannot be computed before $w_t$ has been drawn, since $w_t$ is part of
the context on which $z_{t+1}$ depends. Generation therefore requires one
forward pass \emph{per token}, $T$ passes for $T$ tokens.

\section{On the probabilistic nature of LLMs}
\label{section-summary}

The probabilistic nature of Gen AI is multi-tiered: first, it stems fundamentally from the underlying probabilistic language model, 
which maps semantic properties to joint probability distributions. Second, it is further shaped at the operational level by stochastic techniques and sampling applied during text generation. More precisely:

\medskip
\noindent\textbf{(1) The object learned is a probability measure.}
The basic objects of the model are probability measures on the space $\Omega$ of finite sequences of words.
It leads to a family of Markov kernels $w_{<t}\mapsto\Ptheta(\cdot\mid w_{<t})$, which by Section~\ref{section-model} is used to predict the next word $w_t$ given the set of word generated until $t-1$: $w_{<t}=\{w_1, \ldots, w_{t-1}\}$. 
The reason is not computational convenience but the nature of language:
for a given context several continuations are simultaneously correct, and a
measure is the only object able to represent a set of admissible answers
together with their relative plausibility. A deterministic map would have to
choose arbitrarily among them and would be penalized for whichever it did not
choose.

\medskip
\noindent\textbf{(2) The estimation is statistical inference.}
The corpus is a finite sample from an unknown $\Pstar$; the criterion
\eqref{eq:klmin} is maximum likelihood, i.e.\ an information projection of
$\Pemp$ onto $\mathcal{P}_\Theta$; the algorithm is stochastic approximation.
The quality of fit is measured by \emph{perplexity},
\begin{equation}
\label{perplexity}
  \mathrm{PP}=\exp\!\left(-\frac{1}{T}\sum_{t=1}^{T}\log\Phat(w_t\mid w_{<t})\right),
\end{equation}
the exponential of the empirical cross-entropy, interpretable as an effective
branching factor. As the entropy, (\ref{perplexity}) means that when the probability $\Phat(.\mid w_{<t}))$ becomes dispersed (i.e. more random) the perplexity becomes higher: the $\mathrm{PP}=10$ means the model is on average as uncertain as
if choosing uniformly among ten tokens. Were the language a stationary ergodic
source of entropy rate $h$, no model could achieve $\mathrm{PP}<e^{h}$
\cite{cover2006}.

\medskip
\noindent\textbf{(3) The generation is a random draw.}
The output is a realization of the Markov chain of Remark~\ref{rem:markov},
started at the prompt and sampled by the rules of
Section~\ref{section-decoding}. Two identical prompts yield different outputs
because two independent simulations of the same chain differ --- not because the
model ``reasons differently''. Temperature and top-$p$ are respectively an
exponential tilting and a conditioning of the transition kernel: they modify the
law of the chain being simulated.
\\
\noindent
Familiar behaviors follow as corollaries. \emph{Non-determinism} is (3).
\emph{Sensitivity to the prompt} is the dependence of a Markov chain on its
initial condition. \emph{Hallucination} is the conjunction of (1) with the
mass-covering property of forward KL: a false but linguistically plausible
statement is a high-$\Phat$ string, and nothing in \eqref{eq:klmin}
distinguishes truth from plausibility --- there is no term in the objective
referring to the world or the truth. This is more than an informal argument: Kalai and
Vempala \cite{kalai2024} prove a statistical lower bound on the hallucination
rate of a calibrated model, of the order of the fraction of facts appearing
exactly once in the training corpus --- a Good--Turing missing-mass estimate ---
independently of the architecture or of the quality of the data.
\noindent
Related to this, $\Phat$ is a measure \emph{on all possible strings}, not only the meaningful ones; $\Phat(w_t\mid w_{<t})=0.7$ is an
estimated token frequency, not a degree of belief in a proposition, and is not
automatically calibrated as one. Post-training with human feedback deforms
$\Phat$ towards a reward $r$ \cite{InstructGPT} via
\begin{equation}
  \max_{\theta}\;\E_{y\sim\pi_\theta(\cdot\mid x)}\!\left[r(x,y)\right]
  -\beta\,D \!\left(\pi_\theta(\cdot\mid x)\,\|\,\pi_{\mathrm{ref}}(\cdot\mid x)\right),
\end{equation}
whose solution $\pi^{*}(y\mid x)\propto\pi_{\mathrm{ref}}(y\mid
x)\exp\{r(x,y)/\beta\}$ is again a Gibbs measure --- a deformation of a
probability law, not a departure from the probabilistic framework.

\begin{remark}[Calibration]
A predictor is \emph{calibrated} when its announced probabilities match observed
frequencies: among all the occasions on which it announces $q$, the event should
occur a fraction $q$ of the time,
\begin{equation}
  \Prob\!\left(\text{event}\;\middle|\;\hat p=q\right)=q
  \qquad\text{for all } q\in[0,1].
\end{equation}
Three observations are in order. First, the training criterion pushes towards calibration:
the log-loss is a strictly proper scoring rule \cite{gneiting2007}, minimized by
the true conditional law. Second, and decisively, the event being calibrated is
\emph{``this token comes next''}, not \emph{``this assertion is true''}. These
are different events, and calibration on the first entails nothing about the
second: a model may be perfectly calibrated on token frequencies while assigning
high probability to a false statement, because the statement is a plausible
continuation. A high $\Phat$ reports agreement with the corpus, not agreement
with the reality or the truth. Third, post-training degrades even token-level calibration:
optimizing a reward concentrates the measure on preferred responses, sharpening
it without a corresponding gain in accuracy, so that an aligned model becomes overconfident, in the sense that
it announces a probability that is higher than the frequency with which it would be in fact correct.
\end{remark}

\paragraph{Generative AI and its predecessors.} It is tempting to summarize the
development by saying that classical neural networks are deterministic and
generative models probabilistic. This is partially true. 
The first half is right, and holds of both: a network --- RNN or
transformer --- is a deterministic map, and given a context and $\theta$ the
logits $z_t$ are entirely determined. What is probabilistic is not the
computation but its \emph{output}, since $z_t$ is read through a softmax as a
measure on $\V$; and this too is common to both, an RNN language model
\cite{mikolov2010} having the same terminal softmax, the same factorization
\eqref{eq:chain} and the same loss \eqref{eq:loss} as a transformer. The two
architectures differ in how the context is summarized --- recurrence against
attention --- not in the nature of what they output. Attention
\cite{bahdanau2014} was itself introduced within sequence-to-sequence models
that were already conditional language models \cite{cho2014,sutskever2014}.
The shift lies in the following places:

\medskip
\emph{The status of the measure.} In discriminative usage, the distribution is a
means: one reads off its mode and discards the rest, so only the ranking of the
top candidates matters. In generative usage, the distribution is the product: one
\emph{samples} from it, and the whole shape of the measure matters, its tail
included --- which is why in Sections~\ref{section-learning}
and~\ref{section-decoding} we were led to maintain consideration of that tail.

\medskip
\emph{The output space.} One passes from a measure on a small label set to a
measure on $\V^{*}$, an unbounded combinatorial space. It is precisely this that 
makes the autoregressive factorization necessary
rather than optional. A classifier over a thousand labels can be specified
exhaustively: one lists the thousand probabilities and the measure is fully
described. A language model admits no such list. With $|\V|\approx5\times10^{4}$
there are roughly $10^{470}$ possible texts of a hundred tokens (a huge number).

\medskip
The measure is therefore specified indirectly. At each position the model
produces only the conditional law of the next token, a vector of
$|\V|\approx5\times10^{4}$ numbers; a hundred such vectors, some $5\times10^{6}$
numbers in all, determine the probability of every one of the $10^{470}$
sequences through the chain rule \eqref{eq:chain}. The factorization is thus not
a modeling convenience but the only possibility by which a measure on this space can
be written.

\emph{The feedback loop.} The output is appended to the input. Hence, the model's own
randomness enters the conditioning of its future draws, and the object
produced is not a single conditional law but a trajectory of a stochastic
process --- the Markov chain of Remark~\ref{rem:markov}. This is what has no
counterpart in a classifier, and it is the reason the whole apparatus of
simulation, and not merely of estimation, is required.
The reorientation is thus from function approximation to \emph{the simulation of
a stochastic process}, the estimation of a conditional law being the common
ground rather than the novelty.

\section{The Diffusion Models Approach : From Gaussian Transitions to Stochastic Differential Equations}
\label{section-diffusion}

Diffusion models provide another probabilistic paradigm for generative
artificial intelligence. While autoregressive language models generate a sequence
of tokens successively by sampling from conditional probability distributions,
diffusion models formulate generation differently: noise is added progressively
to a data point until it becomes indistinguishable from pure Gaussian noise; a
network is then trained to reverse this process step by step, so that, once
trained, it can transform a fresh sample of pure noise into a new, realistic
data point. The training itself only ever sees noise added to real data; it is
at generation time that the process is run in reverse, starting from noise
that has no data underneath it. See the example below.
This approach has become
successful for image generation and has subsequently been extended to audio,
video, three-dimensional data, and scientific and medical imaging
\cite{sohl2015,ho2020,rombach2022}.

\subsection{Object and Motivations}

The goal of generative modeling is to learn an approximation of an
unknown data distribution $p_{\text{data}}(\mathbf{x})$ defined over a
high-dimensional space, and then to draw new samples
$\mathbf{x} \sim p_{\text{data}}$. Here, the symbol $\mathbf{x} \in \mathbb{R}^d$ denotes a single data point (for instance, an
image flattened into a vector of $d$ pixel intensities), and
$p_{\text{data}}$ is the probability density function we wish to model.
Classical approaches---variational autoencoders, generative adversarial
networks, and normalizing flows---try to solve this problem differently but
face the same fundamental difficulty: directly parameterizing and
sampling from a highly complex, multimodal distribution in high dimension is
intractable.

Diffusion models rely on the following idea, which is adapted to contents like images:
instead of learning to map noise to data in one shot, they learn to \emph{gradually}
transform a simple distribution (a standard Gaussian) into the data
distribution through a sequence of small, invertible-in-distribution steps.
This idea was introduced by Sohl-Dickstein et al.~\cite{sohl2015}
and developed by Ho et al.~\cite{ho2020} with Denoising
Diffusion Probabilistic Models (DDPM), and by Song et
al.~\cite{song2021}, who showed that the discrete construction has a
continuous-time limit expressed as a stochastic differential equation (SDE).
The remainder of this section follows this order: we
begin with the discrete-time formulation and the crucial role played by
Gaussian transition kernels, and at the end we take the
continuous-time limit that yields the SDE formulation.

\subsection{The Intuitive Idea}

Let us consider a clean image $\mathbf{x}_0$ and add a small amount of
Gaussian noise to it, over and over again, for $T$ steps. After many steps, the image becomes
unrecognizable before being statistically indistinguishable
from pure Gaussian noise. This is the {\it forward process}: it is fixed and
requires no learning. Now we reverse this operation by starting from pure noise and removing a tiny bit of noise
at each step. If we knew exactly how the noise was added, we could invert the
process and recover a clean image. The {\it reverse process} is what we
learn: a network is trained to undo one step of noising at a time.
Once this training is done on a big data set, we can generate new data by starting from a fresh Gaussian
sample and applying the learned reverse steps. The whole construction is a
probabilistic analogue of annealing: we destroy structure gradually, then
learn to rebuild it gradually.

As an example, suppose we have a dataset of $500$ cat images. We first run the
forward process on each of them: every cat image is progressively randomized
by Gaussian noise until, after $T$ steps, nothing remains but noise. We then
run the learning process: a network is trained to perform the reverse
operation, i.e.\ to remove noise step by step. Once training is complete, we
can use the model to generate a brand-new image of a cat that never existed:
we start from a fresh Gaussian noise sample and apply the learned reverse
process. Moreover, if we feed the network additional information---such as a
text caption---at every denoising step, the model can learn to generate
images conditioned on text, which is the basis of text-to-image systems (see
Section~\ref{subsec:text2img}).

It is worth being precise about what the symbols
$\mathbf{x}_0, \mathbf{x}_1, \dots, \mathbf{x}_T$ represent. Each
$\mathbf{x}_t$ encodes an image as a vector of pixel intensities:
\[
    \mathbf{x}_t = (y_1, y_2, \dots, y_d),
\]
where $y_j$ is the intensity (or color value) at pixel $j$, and $d$ is the
total number of pixels in the image. For a $1024 \times 1024$ RGB image, for
instance, $d = 1024 \times 1024 \times 3 \approx 3.1$ million. So $d$ can be
several million, which is precisely what makes direct density modeling so
difficult and what makes the sequential, Gaussian-based approach of
diffusion models so attractive.

\subsection{Discrete-Time Formulation: Gaussian Transition Probabilities}

\paragraph{Notation.} Let us fix the notation that will be used in the sequel.

\begin{itemize}
    \item $\mathbf{x}_0 \in \mathbb{R}^d$ denotes a clean data point sampled
    from the unknown data distribution $p_{\text{data}}$. The subscript $0$
    indicates ``time zero,'' i.e.\ the clean, uncorrupted data.

    \item $\mathbf{x}_1, \mathbf{x}_2, \dots, \mathbf{x}_T$ denote
    successively noisier versions of $\mathbf{x}_0$. Each
    $\mathbf{x}_t \in \mathbb{R}^d$ has the same dimension as the data. The
    integer $t \in \{0, 1, \dots, T\}$ is a discrete time index, and $T$ is
    the total number of diffusion steps (e.g., $T = 1000$ in image
    applications).

    \item $q(\cdot)$ denotes the \emph{forward} (noising) distribution, which
    is fixed and known.

    \item $p_\theta(\cdot)$ denotes the \emph{reverse} (denoising)
    distribution, parameterized by a neural network with weights $\theta$.

    \item $\mathcal{N}(\boldsymbol{\mu}, \boldsymbol{\Sigma})$ denotes a
    multivariate Gaussian distribution with mean vector
    $\boldsymbol{\mu}$ and covariance matrix $\boldsymbol{\Sigma}$. When
    $\boldsymbol{\Sigma} = \sigma^2 \mathbf{I}$ is isotropic, we write
    $\mathcal{N}(\boldsymbol{\mu}, \sigma^2 \mathbf{I})$. 
		Here $\mathbf{I} \in \mathbb{R}^{d \times d}$ is the identity matrix.

    \item $\beta_1, \dots, \beta_T \in (0, 1)$ is a sequence of small
    positive scalars called the \textbf{noise schedule} (or variance
    schedule). It controls how much noise is injected at each step.
\end{itemize}
\noindent
{\it Remark.} The long form $\mathcal{N}(\mathbf{x}; \boldsymbol{\mu}, \boldsymbol{\Sigma})$ 
gives the \emph{density} of that distribution evaluated at the point $\mathbf{x}$; the
semicolon separates the variable $\mathbf{x}$ from the parameters
$\boldsymbol{\mu}$ and $\boldsymbol{\Sigma}$. Explicitly,
\[
    \mathcal{N}(\mathbf{x}; \boldsymbol{\mu}, \boldsymbol{\Sigma})
    = \frac{1}{(2\pi)^{d/2}|\boldsymbol{\Sigma}|^{1/2}}
      \exp\!\left(
        -\tfrac{1}{2}(\mathbf{x} - \boldsymbol{\mu})^\top
        \boldsymbol{\Sigma}^{-1} (\mathbf{x} - \boldsymbol{\mu})
      \right),
\]
which is a scalar (a density value), whereas
$\mathcal{N}(\boldsymbol{\mu}, \boldsymbol{\Sigma})$ is a distribution. In
what follows we use the long form when we want to be explicit about the
variable---especially for conditional densities such as
$q(\mathbf{x}_t \mid \mathbf{x}_{t-1})$---and the short form when we only
want to name the law.

\subsubsection{The Forward Process}

The forward process is a Markov chain defined by Gaussian transition kernels:
\begin{equation}
    q(\mathbf{x}_t \mid \mathbf{x}_{t-1})
    = \mathcal{N}\!\left(
        \mathbf{x}_t;\ \sqrt{1 - \beta_t}\,\mathbf{x}_{t-1},\ \beta_t \mathbf{I}
      \right),
    \qquad t = 1, \dots, T.
    \label{eq:forward-kernel}
\end{equation}

Let us now examine this expression. The notation
$q(\mathbf{x}_t \mid \mathbf{x}_{t-1})$ denotes the conditional probability
density of the noisy variable $\mathbf{x}_t$ given the previous variable
$\mathbf{x}_{t-1}$. The right-hand side says that $\mathbf{x}_t$ is drawn
from a Gaussian whose mean is a scaled copy of $\mathbf{x}_{t-1}$---namely
$\sqrt{1 - \beta_t}\,\mathbf{x}_{t-1}$---and whose covariance is
$\beta_t \mathbf{I}$, i.e.\ isotropic noise with variance $\beta_t$ in every
coordinate. In the notation
$\mathcal{N}(\mathbf{x}_t; \sqrt{1-\beta_t}\,\mathbf{x}_{t-1}, \beta_t
\mathbf{I})$, the first argument $\mathbf{x}_t$ is the variable, the second
is the mean (which depends on $\mathbf{x}_{t-1}$, hence the conditional
nature of the kernel), and the third is the covariance.

An equivalent and often more convenient way to write the same transition is
via the \emph{reparameterization trick}:
\begin{equation}
    \mathbf{x}_t = \sqrt{1 - \beta_t}\,\mathbf{x}_{t-1}
                 + \sqrt{\beta_t}\,\boldsymbol{\epsilon}_t,
    \qquad \boldsymbol{\epsilon}_t \sim \mathcal{N}(\mathbf{0}, \mathbf{I}).
    \label{eq:reparam}
\end{equation}

Here $\boldsymbol{\epsilon}_t$ is a standard Gaussian noise vector (mean
$\mathbf{0} \in \mathbb{R}^d$, covariance $\mathbf{I}$), and the two
coefficients $\sqrt{1 - \beta_t}$ and $\sqrt{\beta_t}$ are chosen so that
the \emph{variance is preserved}:
\[
    \operatorname{Var}(\mathbf{x}_t)
    = (1 - \beta_t)\,\operatorname{Var}(\mathbf{x}_{t-1}) + \beta_t.
\]
If $\mathbf{x}_{t-1}$ has unit variance, then $\mathbf{x}_t$ also has unit
variance. This {\it variance-preserving} property ensures that the signal
does not blow up or vanish as $t$ grows, and it is why the mean is scaled by
$\sqrt{1 - \beta_t}$ rather than by $1$.

\paragraph{The Gaussian transitions choice.} It is motivated by the universal character of Gaussian distributions which provides
several features and advantages :
\\
\\
\noindent
{\it Closed-form marginals.}
Because the composition of Gaussian transitions is again Gaussian, we can 
write the distribution of $\mathbf{x}_t$ conditioned on the original data
$\mathbf{x}_0$ in closed form. Defining $\alpha_t = 1 - \beta_t$ and
$\bar{\alpha}_t = \prod_{s=1}^{t} \alpha_s$, one obtains
\begin{equation}
    q(\mathbf{x}_t \mid \mathbf{x}_0)
    = \mathcal{N}\!\left(
        \mathbf{x}_t;\ \sqrt{\bar{\alpha}_t}\,\mathbf{x}_0,\
        (1 - \bar{\alpha}_t)\,\mathbf{I}
      \right),
    \label{eq:marginal}
\end{equation}
or equivalently, in reparameterized form,
\begin{equation}
    \mathbf{x}_t = \sqrt{\bar{\alpha}_t}\,\mathbf{x}_0
                 + \sqrt{1 - \bar{\alpha}_t}\,\boldsymbol{\epsilon},
    \qquad \boldsymbol{\epsilon} \sim \mathcal{N}(\mathbf{0}, \mathbf{I}).
    \label{eq:marginal-reparam}
\end{equation}
This single-shot formula is essential: it allows us to sample $\mathbf{x}_t$
directly from $\mathbf{x}_0$ without simulating all intermediate steps,
which makes training efficient. As $t \to T$ and $\bar{\alpha}_t \to 0$, the
marginal $q(\mathbf{x}_t \mid \mathbf{x}_0)$ converges to
$\mathcal{N}(\mathbf{0}, \mathbf{I})$, so the forward process indeed destroys
all information about the data.
\\
\noindent
{\it The central limit theorem.} Even if the individual transitions were not Gaussian, the cumulative effect
of many small independent perturbations tends toward a Gaussian by the
central limit theorem. Choosing Gaussian transitions from the start makes
the limiting distribution exactly Gaussian and analytically tractable.
\\
\\
\noindent
{\it Tractable reverse process.} The reverse transition $q(\mathbf{x}_{t-1} \mid \mathbf{x}_t)$ is also
Gaussian when the forward transitions are Gaussian and $\beta_t$ is small.
Specifically, a standard computation using Bayes' rule gives
\begin{equation}
    q(\mathbf{x}_{t-1} \mid \mathbf{x}_t, \mathbf{x}_0)
    = \mathcal{N}\!\left(
        \mathbf{x}_{t-1};\
        \tilde{\boldsymbol{\mu}}_t(\mathbf{x}_t, \mathbf{x}_0),\
        \tilde{\beta}_t \mathbf{I}
      \right),
    \label{eq:reverse-kernel}
\end{equation}
with
\begin{align}
    \tilde{\boldsymbol{\mu}}_t(\mathbf{x}_t, \mathbf{x}_0)
    &= \frac{\sqrt{\bar{\alpha}_{t-1}}\,\beta_t}{1 - \bar{\alpha}_t}\,\mathbf{x}_0
     + \frac{\sqrt{\alpha_t}\,(1 - \bar{\alpha}_{t-1})}{1 - \bar{\alpha}_t}\,\mathbf{x}_t,
    \label{eq:reverse-mean}\\[4pt]
    \tilde{\beta}_t
    &= \frac{1 - \bar{\alpha}_{t-1}}{1 - \bar{\alpha}_t}\,\beta_t.
    \label{eq:reverse-var}
\end{align}
This closed-form reverse kernel is what makes the training objective simple
and the sampling procedure well-defined.
\\
\\
\noindent
{\it Connection to score matching and Langevin dynamics.} A Gaussian perturbation kernel $q(\mathbf{x}_t \mid \mathbf{x}_0)$ has a
score function $\nabla_{\mathbf{x}_t} \log q(\mathbf{x}_t \mid \mathbf{x}_0)$
that is linear in $\mathbf{x}_t$ (see the digression below for the interest and usage of score functions). This linearity is what allows denoising
score matching to reduce to simple noise prediction, and it is the discrete
precursor of the score-based formulation that later becomes an SDE.

\paragraph{Digression on the Score Function}
\label{rem:score}
For a probability density $p(\mathbf{x})$ on $\mathbb{R}^d$, the
\emph{score function} is defined as the gradient of the logarithm of the
density with respect to the variable,
\begin{equation}
    s(\mathbf{x})
    = \nabla_{\mathbf{x}} \log p(\mathbf{x})
    = \left(
        \frac{\partial \log p}{\partial x_1}(\mathbf{x}),\ \dots,\
        \frac{\partial \log p}{\partial x_d}(\mathbf{x})
      \right) \in \mathbb{R}^d.
    \label{eq:app-score-def}
\end{equation}
It is a vector field : Two properties account for its role in generative modeling.

First, the score is independent of the normalizing constant. If
$p(\mathbf{x}) = \tilde{p}(\mathbf{x}) / Z$ with
$Z = \int \tilde{p}(\mathbf{x})\,d\mathbf{x}$, then
$\nabla_{\mathbf{x}} \log p(\mathbf{x})
= \nabla_{\mathbf{x}} \log \tilde{p}(\mathbf{x})$, since $Z$ does not depend
on $\mathbf{x}$. The score is therefore accessible even when the density
itself is intractable, as in energy-based models where
$p(\mathbf{x}) \propto e^{-E(\mathbf{x})}$.

Second, the score suffices to generate samples via Langevin dynamics,
\begin{equation}
    \mathbf{x}_{k+1}
    = \mathbf{x}_k + \varepsilon\, \nabla_{\mathbf{x}} \log p(\mathbf{x}_k)
      + \sqrt{2\varepsilon}\,\mathbf{z}_k,
    \qquad \mathbf{z}_k \sim \mathcal{N}(\mathbf{0}, \mathbf{I}),
    \label{eq:app-langevin}
\end{equation}
where the gradient term drives samples toward high-density regions and the
injected noise prevents collapse onto a single mode. Equation
\eqref{eq:app-langevin} is the discrete counterpart of the overdamped
Langevin stochastic differential equation, which constitutes the elementary
building block of score-based generative models.

Within the diffusion framework, the relevant quantity is the score of the
Gaussian transition kernel. From the closed-form marginal
$q(\mathbf{x}_t \mid \mathbf{x}_0)
= \mathcal{N}(\mathbf{x}_t; \sqrt{\bar{\alpha}_t}\,\mathbf{x}_0,
(1 - \bar{\alpha}_t)\mathbf{I})$, differentiation of the log-density with
respect to $\mathbf{x}_t$, combined with the reparameterization
$\mathbf{x}_t - \sqrt{\bar{\alpha}_t}\,\mathbf{x}_0
= \sqrt{1 - \bar{\alpha}_t}\,\boldsymbol{\epsilon}$, yields
\begin{equation}
    \nabla_{\mathbf{x}_t} \log q(\mathbf{x}_t \mid \mathbf{x}_0)
    = -\frac{\boldsymbol{\epsilon}}{\sqrt{1 - \bar{\alpha}_t}}.
    \label{eq:app-score-noise}
\end{equation}
Thus, up to the scalar factor $-1/\sqrt{1-\bar{\alpha}_t}$, the score of the
conditional Gaussian coincides with the noise $\boldsymbol{\epsilon}$ that
was added. Estimating the noise is therefore equivalent to estimating the
score. When the network $\boldsymbol{\epsilon}_\theta(\mathbf{x}_t, t)$ is
trained to predict $\boldsymbol{\epsilon}$, it implicitly learns
$s_\theta(\mathbf{x}_t, t) \approx
-\boldsymbol{\epsilon}_\theta(\mathbf{x}_t, t)/\sqrt{1 - \bar{\alpha}_t}$,
where the marginal score $\nabla_{\mathbf{x}_t} \log q(\mathbf{x}_t)$ is
recovered from the conditional one through the denoising score matching
identity, also known as Tweedie's formula,
\begin{equation}
    \nabla_{\mathbf{x}_t} \log q(\mathbf{x}_t)
    = \mathbb{E}_{q(\mathbf{x}_0 \mid \mathbf{x}_t)}
      \big[ \nabla_{\mathbf{x}_t} \log q(\mathbf{x}_t \mid \mathbf{x}_0) \big].
    \label{eq:app-tweedie}
\end{equation}
This identity justifies the substitution of the intractable marginal score
by a regression on the tractable conditional score and explains the
explicit appearance of the score in the reverse SDE in the stochastic differential equation approach (see below), whose
drift is $f(\mathbf{X}_t, t) - g(t)^2 \nabla_{\mathbf{x}} \log
p_t(\mathbf{X}_t)$.

\subsubsection{The Reverse Process and the Training Objective}

Since the true reverse kernel $q(\mathbf{x}_{t-1} \mid \mathbf{x}_t)$ depends
on the unknown data distribution, we approximate it with a learned Gaussian:
\begin{equation}
    p_\theta(\mathbf{x}_{t-1} \mid \mathbf{x}_t)
    = \mathcal{N}\!\left(
        \mathbf{x}_{t-1};\
        \boldsymbol{\mu}_\theta(\mathbf{x}_t, t),\
        \sigma_t^2 \mathbf{I}
      \right),
    \label{eq:learned-reverse}
\end{equation}
where $\boldsymbol{\mu}_\theta(\mathbf{x}_t, t)$ is the output of a
network with weights $\theta$, and $\sigma_t^2$ is a fixed or learned
variance. Training maximizes the variational lower bound on
$\log p_\theta(\mathbf{x}_0)$, which after simplification
\cite{ho2020} reduces to a mean-squared error on the noise:
\begin{equation}
    \mathcal{L}_{\text{simple}}(\theta)
    = \mathbb{E}_{t, \mathbf{x}_0, \boldsymbol{\epsilon}}
      \left[
        \left\|
          \boldsymbol{\epsilon}
          - \boldsymbol{\epsilon}_\theta\!\left(
              \sqrt{\bar{\alpha}_t}\,\mathbf{x}_0
              + \sqrt{1 - \bar{\alpha}_t}\,\boldsymbol{\epsilon},\ t
            \right)
        \right\|^2
      \right],
    \label{eq:ddpm-loss}
\end{equation}
where $\boldsymbol{\epsilon} \sim \mathcal{N}(\mathbf{0}, \mathbf{I})$ is the
noise used to form $\mathbf{x}_t$, and $\boldsymbol{\epsilon}_\theta$ is the
network's prediction of that noise. Intuitively, the network learns to look
at a noisy image $\mathbf{x}_t$ and guess which noise vector was added---%
equivalently, to estimate the score
$\nabla_{\mathbf{x}_t} \log q(\mathbf{x}_t)$ up to a scaling factor ; we refer to
\cite{ho2020} for the calculation details.

\subsection{From Unconditional Generation to Text-to-Image}
\label{subsec:text2img}

\subsubsection{Unconditional generation}

Start with the unconditional case. We train a network
$\boldsymbol{\epsilon}_\theta(\mathbf{x}_t, t)$ to predict the noise added
at step $t$. Once trained, generation proceeds as follows:
\begin{enumerate}
    \item Sample $\mathbf{x}_T \sim \mathcal{N}(\mathbf{0}, \mathbf{I})$.
    This is a $d$-dimensional vector of pure Gaussian noise---for an image,
    think of a $512 \times 512 \times 3$ tensor of independent Gaussian
    pixels. Visually, it looks like television static.

    \item For $t = T, T-1, \dots, 1$, compute the predicted noise
    $\boldsymbol{\epsilon}_\theta(\mathbf{x}_t, t)$ and use it to form an
    estimate of $\mathbf{x}_{t-1}$ according to the reverse Gaussian kernel.

    \item After $T$ steps, we get the output $\mathbf{x}_0$.
\end{enumerate}
The fact is that $\mathbf{x}_0$ is a \emph{new sample} from (an
approximation of) $p_{\text{data}}$. If the model was trained on a dataset
of cat photos, $\mathbf{x}_0$ will look like a cat that never existed. If it
was trained on handwritten digits, $\mathbf{x}_0$ will look like a digit.
The Gaussian noise is progressively sculpted into a structured image.

But this unconditional process gives us no control: we cannot ask for a cat
specifically, or a cat wearing a hat. That control comes from {\it conditioning}.

\subsubsection{Conditioning on text}

In text-to-image generation, we want to sample from the conditional
distribution $p(\mathbf{x} \mid \mathbf{c})$, where $\mathbf{c}$ is a text
prompt such as ``a photo of futuristic cars'' The prompt is
first encoded into a vector (or a sequence of vectors) $\mathbf{c}$ by a
text encoder---for example CLIP's text encoder or a T5 encoder. The denoising
network then takes both the noisy image $\mathbf{x}_t$ and the text embedding
$\mathbf{c}$ as inputs:
\[
    \boldsymbol{\epsilon}_\theta(\mathbf{x}_t, t, \mathbf{c}).
\]
Training uses pairs $(\mathbf{x}_0, \mathbf{c})$ of images and their
captions. At each step, noise $\boldsymbol{\epsilon}$ is added to
$\mathbf{x}_0$ to form $\mathbf{x}_t$, and the network is trained to predict
$\boldsymbol{\epsilon}$ given $\mathbf{x}_t$, $t$, and $\mathbf{c}$. The loss
is the same mean-squared error as before, but now the network sees the text
as an additional input.

At generation time, the procedure is identical to the unconditional case,
except that at every denoising step the network is told the prompt
$\mathbf{c}$. Concretely:
\begin{enumerate}
    \item Encode the prompt ``a photo of futuristic cars'' into
    $\mathbf{c}$.
    \item Sample $\mathbf{x}_T \sim \mathcal{N}(\mathbf{0}, \mathbf{I})$.
    \item For $t = T, \dots, 1$, predict
    $\boldsymbol{\epsilon}_\theta(\mathbf{x}_t, t, \mathbf{c})$ and update
    $\mathbf{x}_t \to \mathbf{x}_{t-1}$.
    \item Output $\mathbf{x}_0$, which is now a sample from
    $p(\mathbf{x} \mid \mathbf{c})$.
\end{enumerate}
The same noise seed combined with different prompts yields different images;
the same prompt with different noise seeds yields different images. This is
the mechanism behind {\bf Stable Diffusion, DALL$\cdot$E~2, Imagen, and
Midjourney}.

\subsubsection{Why this works: the score interpretation}

Recall that predicting the noise $\boldsymbol{\epsilon}$ is equivalent (up to
a scaling factor) to estimating the score
$\nabla_{\mathbf{x}_t} \log p_t(\mathbf{x}_t \mid \mathbf{c})$. In the
conditional case, the score points in the direction in which the noisy image
should be modified to become more likely under the text-conditioned
distribution. Each denoising step is a small move along this score, plus a
bit of injected noise to keep the process stochastic. Over many steps, these
small moves accumulate into a coherent image that matches the prompt.

\subsubsection{Classifier-free guidance}

A crucial practical ingredient is \textbf{classifier-free guidance}
\cite{ho2022cfg}. At each step, the network is evaluated twice: once with
the prompt $\mathbf{c}$ and once with an empty prompt $\varnothing$. The two
predictions are combined as
\begin{equation}
    \tilde{\boldsymbol{\epsilon}}_\theta(\mathbf{x}_t, t, \mathbf{c})
    = \boldsymbol{\epsilon}_\theta(\mathbf{x}_t, t, \varnothing)
      + s \left[
          \boldsymbol{\epsilon}_\theta(\mathbf{x}_t, t, \mathbf{c})
          - \boldsymbol{\epsilon}_\theta(\mathbf{x}_t, t, \varnothing)
        \right],
    \label{eq:cfg}
\end{equation}
where $s > 1$ is the guidance scale. This sharpens the alignment between the
generated image and the prompt, at the cost of some diversity. Without it,
text-to-image models tend to produce images that are only loosely related to
the prompt.

\subsubsection{Latent diffusion}

In practice, the denoising is often not performed in pixel space. Stable
Diffusion \cite{rombach2022} uses a {\it latent diffusion} approach: a
variational autoencoder first compresses the image into a lower-dimensional
latent representation
$\mathbf{z}_0 \in \mathbb{R}^{h \times w \times c}$ with $h, w$ much smaller
than the original image dimensions. The diffusion process is then run in this
latent space, and the final latent $\mathbf{z}_0$ is decoded back to pixel
space by the VAE decoder. This makes training and sampling far more
computationally efficient while preserving perceptual quality.

\subsubsection{Summary}

The reverse diffusion process \emph{starts} from noise and \emph{transforms} it into data. 
In text-to-image generation, the transformation is guided at every step by a text embedding, so the final
image is a sample from the conditional distribution
$p(\mathbf{x} \mid \mathbf{c})$. The same mathematical machinery---Gaussian
forward transitions, learned reverse transitions, and score estimation---%
underlies unconditional image generation, text-to-image, image editing,
super-resolution, and many other tasks.

\subsection{Continuous-Time Limit: Stochastic Differential Equations}

The discrete formulation above becomes difficult to manipulate when $T$ is large. Taking the 
limit $T \to \infty$ with $\beta_t \to 0$
yields a continuous-time description in terms of stochastic differential
equations (Song et al.~\cite{song2021}).

Let $t \in [0, T]$ be a continuous time variable,
let $\mathbf{X}_t \in \mathbb{R}^d$ be a continuous-time stochastic process,
$f(\mathbf{X}_t, t) \in \mathbb{R}^d$ a drift, $g(t) \in \mathbb{R}$ a
scalar diffusion coefficient, and $\mathbf{B}_t \in \mathbb{R}^d$ a standard
Brownian motion. The forward SDE is
\begin{equation}
    d\mathbf{X}_t = f(\mathbf{X}_t, t)\,dt + g(t)\,d\mathbf{B}_t,
    \qquad \mathbf{X}_0 \sim p_{\text{data}}.
    \label{eq:forward-sde}
\end{equation}
The reverse-time SDE, obtained from Anderson's theorem \cite{anderson1982},
is, with $\overleftarrow{\mathbf{X}}_t$ denoting the reverse process,
\begin{equation}
    d\overleftarrow{\mathbf{X}}_t
    = \left[
        f(\overleftarrow{\mathbf{X}}_t, t)
        - g(t)^2 \nabla_{\mathbf{x}} \log p_t(\overleftarrow{\mathbf{X}}_t)
      \right] dt
      + g(t)\,d\bar{\mathbf{B}}_t,
    \qquad \overleftarrow{\mathbf{X}}_T \sim \mathcal{N}(\mathbf{0}, \mathbf{I}),
    \label{eq:reverse-sde}
\end{equation}
where $p_t$ is the marginal density of $\mathbf{X}_t$ and
$\nabla_{\mathbf{x}} \log p_t$ is its score. The reverse SDE has a
deterministic counterpart, the \emph{probability flow ODE}: Using the It\^o formula, Song et al. showed
that the solution $\mathbf{Y}_t$ of the (deterministic) ODE:
\begin{equation}
    \frac{d\mathbf{Y}_t}{dt}
    = f(\mathbf{Y}_t, t)
      - \frac{1}{2}g(t)^2 \nabla_{\mathbf{x}} \log p_t(\mathbf{Y}_t),
    \qquad \mathbf{Y}_T \sim \mathcal{N}(\mathbf{0}, \mathbf{I}).
    \label{eq:prob-flow-ode}
\end{equation}
 has the same marginal law as that of $\mathbf{X}_t$ at every time $t$: 
the two share the same marginals, but not the same paths. 
It is understood that this ODE is also solved backward with the same condition 
$Y_T=X_T=\mathcal{N}(\mathbf{0}, \mathbf{I})$.
Integrating \eqref{eq:prob-flow-ode} backwards from
$t = T$ to $t = 0$ produces samples from $p_{\text{data}}$, and because the
map is deterministic and invertible, it also yields exact log-likelihoods
via the change-of-variables formula. We omit the details and refer to 
\cite{song2021}, \cite{tang2024}. The discrete analogue of this ODE is the DDIM sampler \cite{song2021ddim}.

\subsection{Current Usage and comments}

Diffusion models have achieved high quality results in image generation
(Stable Diffusion, DALL$\cdot$E~2, Imagen), image editing, super-resolution,
medical imaging, audio and speech synthesis, video generation, reinforcement
learning, and computational biology (protein structure and molecule design).
Their appeal lies in training stability, mode coverage, and the principled
probabilistic framework that connects them to score matching, Langevin
dynamics, and continuous normalizing flows. A recent survey is given by
Ahsan et al.~\cite{ahsan2025}.

The power of diffusion models comes from the probabilistic features of the
Gaussian transitions: they are analytically tractable, compose into closed-form
marginals, and admit a Gaussian reverse kernel that a neural network can
learn. The discrete construction of DDPM is the practical realization of
this insight; the continuous-time SDE formulation is its mathematically
natural limit, unifying discrete and continuous views and linking generative
modeling to the classical theory of stochastic processes. In text-to-image
systems, the same machinery, augmented with a text encoder and
classifier-free guidance, produces images conditioned on natural-language
prompts---one of the most visible successes of generative AI today.


\end{document}